\documentclass[10pt,conference]{IEEEtran}
\usepackage{cite}
\usepackage{amsmath,amssymb,amsfonts}

\usepackage{graphicx}
\usepackage{textcomp}
\usepackage{xcolor}
\usepackage[hyphens]{url}
\usepackage{fancyhdr}
\usepackage{hyperref}
\usepackage{multirow}
\usepackage{array}
\usepackage{algorithm}
\usepackage{algpseudocode}
\usepackage{amsmath}

\newcommand{\hpcayear}{2026}

\newcommand{\hpcasubmissionnumber}{572}
\newcommand{\paper}{RED-PIM}
\title{\paper{}: Reducing Data Movement for Transformers using Processing-in-Memory} 

\def\hpcacameraready{} 

\newcommand\hpcaauthors{Zahra Yousefijamarani, Alaa Alameldeen}
\newcommand\hpcaaffiliation{Simon Fraser University}
\newcommand\hpcaemail{}

\author{
  \ifdefined\hpcacameraready
    \IEEEauthorblockN{\hpcaauthors{}}
      \IEEEauthorblockA{
        \hpcaaffiliation{} \\
        \hpcaemail{}
      }
  \else
    \IEEEauthorblockN{\normalsize{HPCA \hpcayear{} Submission
      \textbf{\#\hpcasubmissionnumber{}}} \\
      \IEEEauthorblockA{
        Confidential Draft \\
        Do NOT Distribute!!
      }
    }
  \fi 
}

\fancypagestyle{camerareadyfirstpage}{%
  \fancyhead{}
  
  \fancyfoot[C]{}
}
\begin{document}
\maketitle

\ifdefined\hpcacameraready 
  \thispagestyle{camerareadyfirstpage}
  \pagestyle{empty}
\else
  \thispagestyle{plain}
  \pagestyle{plain}
\fi

\newcommand{\hpcaheight}{0mm}
\ifdefined\eaopen
\renewcommand{\hpcaheight}{12mm}
\fi

\begin{abstract}

Transformers are widely used across many domains, including natural language processing, computer vision, web search, and DNA sequence analysis. Given their broad applicability, improving the performance of transformer models is critical. However, the high volume of data movement between processing units and memory during attention operations significantly limits their efficiency.
Processing-In-Memory (PIM) mitigates this issue by performing computations directly inside memory. While prior work has proposed PIM-based transformer implementations, they suffer from costly inter-bank communication, and struggle to scale due to the limited capacity of memory banks. As a result, attention-related data must be split across banks, diminishing the potential benefits of PIM.

In this work, we propose \paper{}, an algorithm-architecture co-design that reduces attention latency by minimizing inter-bank data movement from $O(N^2)$ to $O(N)$ and shrinking intermediate attention matrices from $N \times N$ to $d \times d$. By reorganizing matrix operations, performing computations locally, and employing an optimized data transfer strategy, \paper{} significantly reduces computation cost and interconnect traffic. Compared to baseline PIM implementation, \paper{} achieves inference time reductions ranging from 16.05\% to 99.99\% (geometric mean of 66.42\%), with the largest gains on longer sequences. On real-world datasets, \paper{} improves performance by 99.60\% for long documents and 13.44\% for shorter ones, while maintaining or improving accuracy. These results demonstrate \paper{}'s effectiveness for scalable and efficient transformer inference.

\end{abstract}

\section{Introduction}
Traditional CPU and GPU-based architectures face significant challenges in meeting the computational demands of large-scale neural networks due to inefficiencies caused by data movement between memory and processing units. This back-and-forth transfer of massive amounts of data creates bottlenecks that degrade performance and increase energy consumption.

Among various deep learning models, this issue becomes more significant in Transformer-based architectures \cite{vaswani2017attention}, mainly due to their self-attention module.
The reason is that self-attention involves computing attention scores between all pairs of input tokens. To this end, after the initial calculation of the query, key, and value vectors from input tokens, the model has to calculate the similarity between all queries and keys to generate an attention map. This requires $O(N^2)$ data movements for (query, key) comparisons, where $N$ is the sequence length. Subsequently, multiplying this attention map with the value vectors to obtain the final result also involves $O(N^2)$ movements.

Due to the large volume of data needed in large-scale transformers, typical CPU or GPU caches are ineffective. Self-attention computations generate extremely large intermediate data structures that far exceed typical cache capacities, leading to frequent cache evictions and increased latency. To illustrate, consider a sequence with 8,000 tokens. The resulting attention map consists of 64 million values, requiring approximately 32MB of memory even when using the lowest-precision format (i.e., FP16), a much larger size than typical CPU or GPU cache capacities. 

Processing In/Near Memory (PIM) \cite{yang2020retransformer, zhou2022transpim, sridharan2023x, laguna2022hardware, lu2023rram, li2023h3datten, kang2021framework, li2024specpim} offers an alternative solution to reduce the data movement between the memory and processing units.  PIM addresses this challenge by integrating computation directly within memory arrays, thereby eliminating the need for frequent data transfers between the processor(s) and memory.
However, even with PIM, the limited capacity of memory banks remains a major barrier for scaling transformer models. For instance, a typical HBM bank has only 32MB of storage, which supports up to 8,000 tokens in FP16 precision. In contrast, modern large language models (LLMs) such as GPT‑4.1 \cite{gpt41}, Gemini 2 \cite{gemini2}, and Grok‑3 \cite{grok} support much longer sequences, up to 1 million tokens. Attempting to store full attention maps across multiple banks introduces significant inter-bank communication overhead, which undermines the performance and energy-efficiency advantages that PIM is meant to provide.

In this paper, we propose \paper{} that reduces the latency of the attention operation by leveraging PIM and reducing the inter-bank data movement. \paper{} achieves this by adopting an alternative algorithm for attention computation which eliminates the need to construct the full $N \times N$ attention map. Instead, it operates on and stores only a compact $d \times d$ matrix, where $d$ is the latent dimension in the self-attention module and $d \ll N$. Specifically, \paper{} employs efficient attention \cite{shen2021efficient} to significantly reduce memory usage while maintaining model performance. We reorganize the attention operations to keep computations more local, reduce data movement, and create smaller intermediate results that can fit in a single memory bank.

Along with reorganizing the attention operations, we introduce architectural optimizations, including near-bank Processing Control Units (PCUs) that perform local computations with optimized implementations. We propose architectural enhancements to optimize inter-bank data flow via a hierarchical aggregation strategy among memory banks, which results in lower data transfer and latency.

In this paper, we make the following contributions:

\begin{itemize}
    \item \textbf{Bottleneck Analysis of Standard Self-Attention}: We show that the standard Query-Key multiplication in self-attention introduces two key inefficiencies: (1) it causes significant data movement due to the all-to-all token comparisons across the input sequence, and (2) it produces a large intermediate matrix (attention map) that often exceeds the storage capacity of a single HBM bank. As a result, the subsequent Attention-Value multiplication requires additional inter-bank communication to access the full score matrix.
    \item \textbf{Algorithmic Optimizations for Reduced Data Movement}: By restructuring Scaled Dot-Product Attention (SDPA), we optimize data locality, minimizing redundant memory accesses and reducing data movement between HBM banks from $O(N^2)$ to $O(N)$.
    
    \item \textbf{Architectural Design for In-HBM Processing}: We integrate our architecture with FIMDRAM~\cite{kwon202125}, leveraging PIM capabilities to execute self-attention computations directly within memory banks with optimized implementation. A specialized network topology facilitates efficient inter-bank communication. We introduce a hierarchical aggregation strategy that progressively refines attention computations and reduces redundant data transfers between memory banks.
    We design a softmax function optimized for hardware using bitwise operations and lookup tables to accelerate exponentiation. This approach maintains accuracy while reducing computational complexity and memory usage in transformer models.
    
    \item \textbf{Customized Simulation Framework for Data Movement Analysis}: Building on the DAMOV simulator~\cite{oliveira2021damov}, we develop a tailored simulation environment to evaluate and model inter-bank data movement in HBM architectures, enabling detailed analysis of communication patterns for attention mechanisms.
    
    \item \textbf{Extensive Evaluation Demonstrating Significant Performance Gains}: Our evaluation shows inference time reductions ranging from 16.05\% to 99.99\% (geometric mean of 66.42\%) compared to a baseline PIM implementation, with the greatest improvements observed on longer sequences. On real-world datasets, we achieve a geometric mean improvement of 99.56\% for long documents (IMDB, PubMed, Arxiv, WikiHop, and GovReport~\cite{cohan2018discourse, imdb, wikihop, govreport}) and 13.44\% for shorter documents (GLUE benchmark~\cite{wang2018glue}), while maintaining or improving accuracy.
\end{itemize}

In the remainder of this paper, we discuss some background on transformer models, processing-in-memory, and High-Bandwidth Memory in Section~\ref{sec:background}. We present our proposed algorithmic changes and architecture in Section~\ref{sec:method}. We discuss our methodology and results in Section~\ref{sec:evaluation}. Section~\ref{related-work} discusses related work, and we conclude in Section~\ref{sec:conclusion}.


    


\section{Background} 
\label{sec:background}

\subsection{Transformers}
\begin{figure}[t]
\centering
\includegraphics[width=0.49\textwidth]{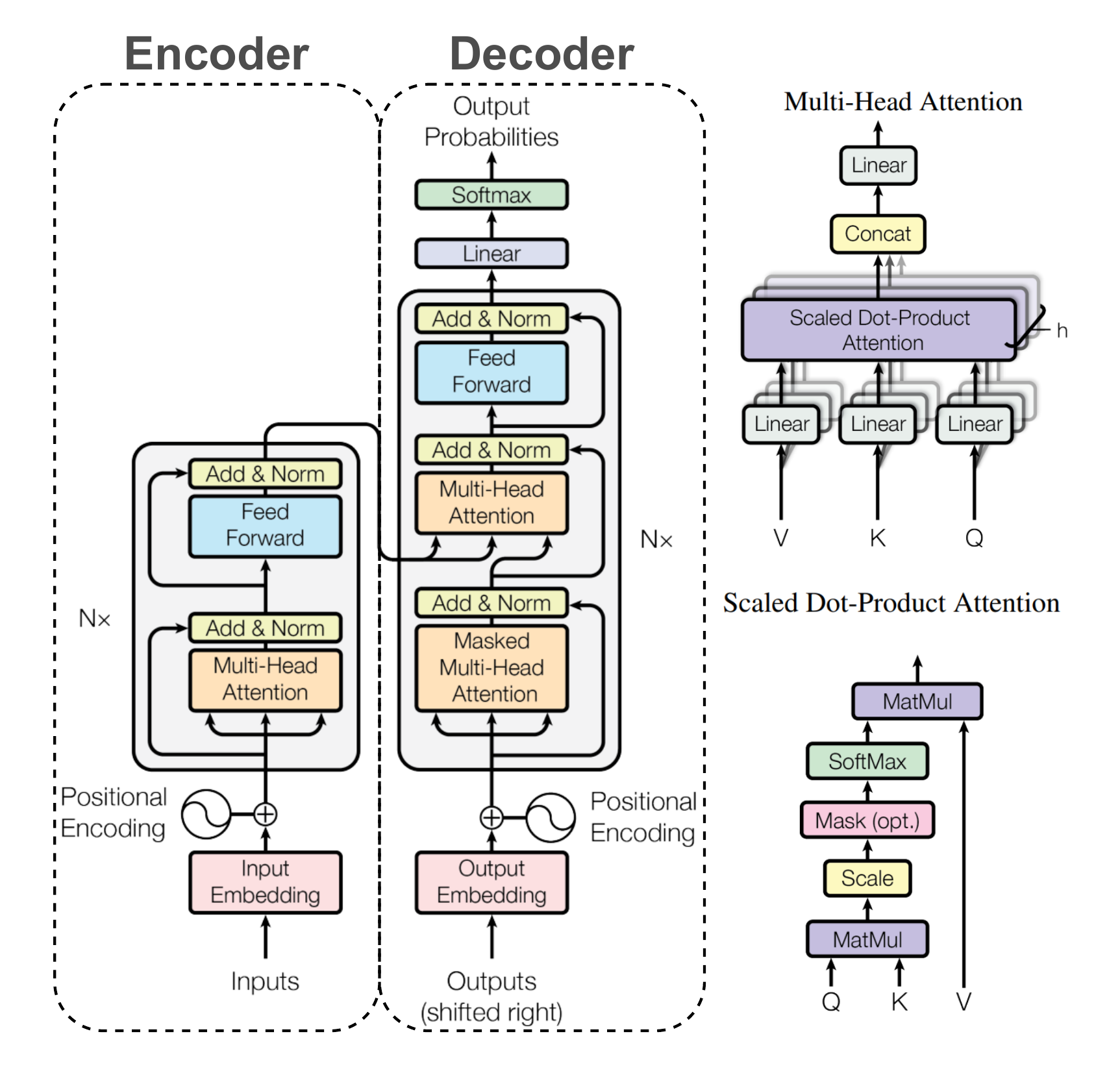}
\caption{The Transformer model architecture \cite{vaswani2017attention}.}
\label{fig:transformer}
\end{figure}

Transformers, introduced by Vaswani et al. in 2017 \cite{vaswani2017attention}, are a class of deep learning models that have become widely used in various applications. Their popularity was a result of their ability to process sequential data in parallel, and effectively capture long-range dependencies. Transformer-based architectures are typically built using a stack of encoder blocks, decoder blocks, or both.
Each block consists of several key components: Multi-Head Attention (MHA), Feed-Forward Neural Networks (FFN), and Layer Normalization, as illustrated in Figure \ref{fig:transformer} \cite{vaswani2017attention}.

The \textit{MHA block} is designed to capture complex relationships between tokens in a sequence. Each attention head computes attention scores that quantify the influence of one token on another. The outputs from multiple attention heads are then combined and processed to generate a comprehensive contextual representation of the input sequence. To generate such representation, 
the MHA block first derives the Query (Q), Key (K), and Value (V) matrices by applying linear transformations to the input sequence $X = [x_1, x_2, ..., x_N]$. These transformations use trainable weight matrices: the Query matrix ($W_Q$), Key matrix ($W_K$), and Value matrix ($W_V$):

\begin{equation}
Q = X.W_Q,\quad V = X.W_V,\quad K = X.W_K
\end{equation}

The Scaled Dot-Product Attention mechanism then computes attention scores as follows:

\begin{equation}
\label{eq:sda}
Attention(K, Q, V) = softmax(\frac{Q.K^T}{\sqrt{d}}). V
\end{equation}

where $d$ is the dimensionality of the key vectors. The scaling factor $\sqrt{d}$ prevents the dot product from becoming too large, which could push the softmax function into regions with very small gradients that would hinder effective learning.
The outputs of multiple attention heads are concatenated and passed through another linear layer to combine their results and generate the final representation.

The \textit{FFN layer} receives the output of the MHA block and transforms it into a more expressive representation. This transformation typically involves applying two linear layers with an intermediate activation function (e.g., ReLU) as follows:

\begin{equation}
 FFN(x) = max(0, x.W_1 + b_1).W_2 + b_2
\end{equation}

\textit{Layer normalization and residual connections} are crucial for enhancing stability and improving training efficiency of transformers. Residual connections bypass the attention and feed-forward layers, while layer normalization ensures consistent input scaling, which improves gradient flow.

In recent years, transformers have grown significantly. Current transformer models contain billions of parameters \cite{wang2024deepnet}. In addition to the size of the model, the sequence length has also increased to accommodate more complex tasks, further amplifying the computational and memory demands of these models \cite{ding2023longnet, dubey2024llama}. Although these increases in size and sequence length have improved performance across various domains, they have also led to significantly longer execution times and higher computational costs.

\subsection{Processing In/Near Memory Architectures}

The concept of PIM has emerged as a response to the increasing computational bottlenecks caused by data movement between memory and processing unit in traditional von Neumann architectures \cite{ghose2019processing}. In these architectures, the separation of memory and compute units leads to significant energy and time costs, especially in data-intensive applications such as machine learning, graph processing, and high-performance computing \cite{ahn2015scalable, boroumand2018google, ghose2019processing}. 

PIM addresses this challenge by integrating computational capabilities directly within the memory arrays, enabling data to be processed close to where it is stored. This approach eliminates the need for frequent data transfers between the processor(s) and memory, significantly reducing both execution time and energy consumption \cite{mutlu2022modern}.


\subsection{Prior PIM-Based Accelerators for Transformers}

Prior research on PIM-based accelerators for Transformers can be broadly classified into two categories. 

The first category comprises hybrid architectures that integrate host-based accelerators with near-memory processors to enhance transformer performance \cite{ding2023haima, heo2024neupims, singh2024dram}. While these approaches aim to combine the advantages of traditional and PIM systems, they often suffer from high communication overhead between the host and memory. Additionally, they introduce substantial architectural complexity, which limits scalability and energy efficiency.

The second category consists of fully PIM-based transformer accelerators, which execute computations entirely within memory. These include both analog PIM designs \cite{yang2020retransformer, sridharan2023x, laguna2022hardware, lu2023rram} and DRAM-based solutions \cite{zhou2022transpim, singh2024dram}. Despite their promise, these designs still face key challenges: They struggle with the large intermediate data sizes generated by self-attention, and inter-bank communication remains a major bottleneck, particularly for long sequences. Furthermore, many analog PIM approaches face limitations related to precision, reliability, and non-ideal analog behavior, which hinder their practicality and accuracy in real-world deployments.

We further discuss these proposals and their limitations in Section~\ref{related-work}.

\subsection{High Bandwidth Memory}
\begin{figure*}[t]  
    \centering
    \includegraphics[width=\textwidth]{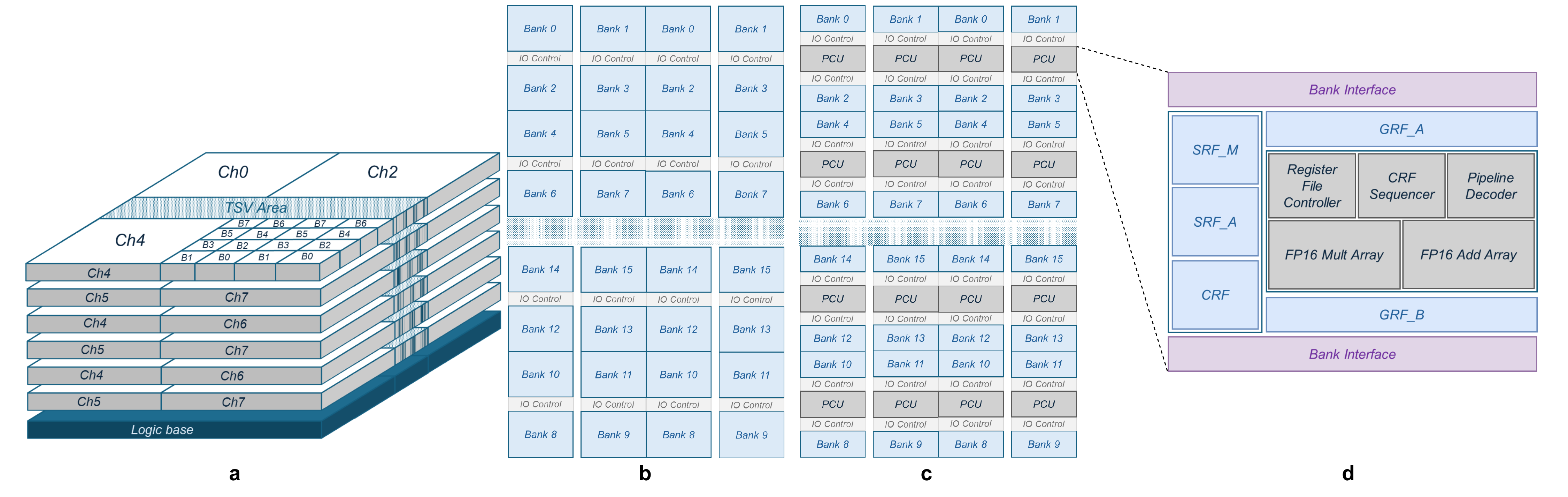}
    \caption{a) HBM architecture; b) Conventional HBM2; c) FIMDRAM architecture; d) Computing unit block.}
    \label{fig:hbm_arch}
\end{figure*}

HBM, illustrated in Figure \ref{fig:hbm_arch}.a, is a type of 3D-stacked DRAM designed to deliver significantly higher memory bandwidth compared to traditional memory technologies like DDR or GDDR. HBM achieves this by vertically stacking multiple memory dies, interconnecting them using through-silicon vias, and placing the stack close to the processor on the same substrate \cite{sohn20161}. This architecture provides higher capacity, reduced latency, and increased effective data bandwidth \cite{jun2017hbm}. HBM is widely employed in high-performance computing (HPC), networking, and graphics applications \cite{zhu2018performance, 9908071, jun2017hbm}, where memory bandwidth and latency are critical factors.

An example of PIM-based HBM is Function-In-Memory DRAM (FIMDRAM) \cite{kwon202125} that was introduced by Samsung in 2021. FIMDRAM is a real PIM prototype implementation based on the HBM2 memory standard (Figure \ref{fig:hbm_arch}.b). FIMDRAM (also known as HBM-PIM) incorporates 16-wide single-instruction multiple-data (SIMD) engines into memory banks, leveraging bank-level parallelism while maintaining the optimized memory subarray. Each pair of memory banks includes a programmable computing unit (PCU) as shown in Figure \ref{fig:hbm_arch}.c, enhancing parallelism and processing bandwidth. By performing computations directly within memory, FIMDRAM significantly reduces data movement, improving efficiency for memory-intensive tasks.

Figure \ref{fig:hbm_arch}.d illustrates the internal architecture of a PCU. Each PCU includes two sets of SIMD floating-point units (FPUs): one dedicated to addition and the other to multiplication. Each set contains 16 FPUs, where each FPU is 16 bits wide.

The PCU is equipped with three kind of registers for managing computations:
\begin{itemize}
    \item \textbf{Command Register File (CRF)} which serves as an instruction buffer capable of holding up to 32 instructions, where each instruction is 32 bits.
    \item \textbf{General Register Files (GRFs)} which consist of 16 registers, each 256-bits wide. These registers are used for intermediate data storage.
    \item \textbf{Scalar Register Files (SRFs)} designed to store up to 16 scalar values, each 16-bits wide.
\end{itemize}

The PCU’s instruction set includes nine 32-bit RISC instructions, categorized into three groups: Control flow instructions: \texttt{NOP}, \texttt{JUMP}, and \texttt{EXIT}; Arithmetic instructions: \texttt{ADD}, \texttt{MUL}, \texttt{MAC}, and \texttt{MAD}; and Data movement instructions: \texttt{MOV} and \texttt{FILL}.

\section{\paper{}: Proposed Design}
\label{sec:method}

\subsection{Baseline Implementation of Attention in PIM Architecture}
\label{base_impl}

\begin{figure}[t]
\centering
\includegraphics[width=0.48\textwidth]{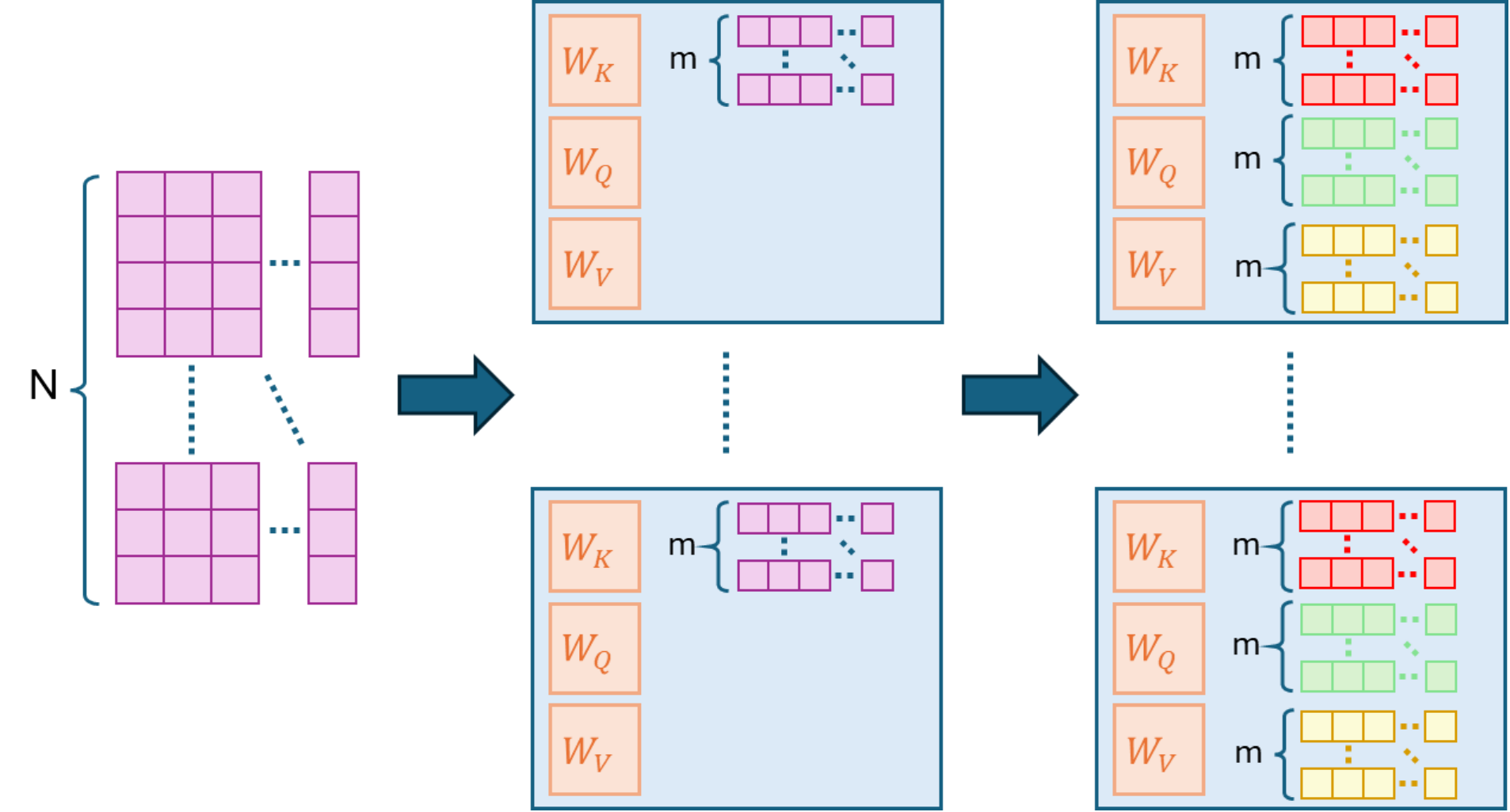}
\caption{Token and weight matrix mapping across banks.}
\label{fig:mapping}
\end{figure}

\begin{figure*}[t]  
    \centering
   \includegraphics[width=\textwidth]{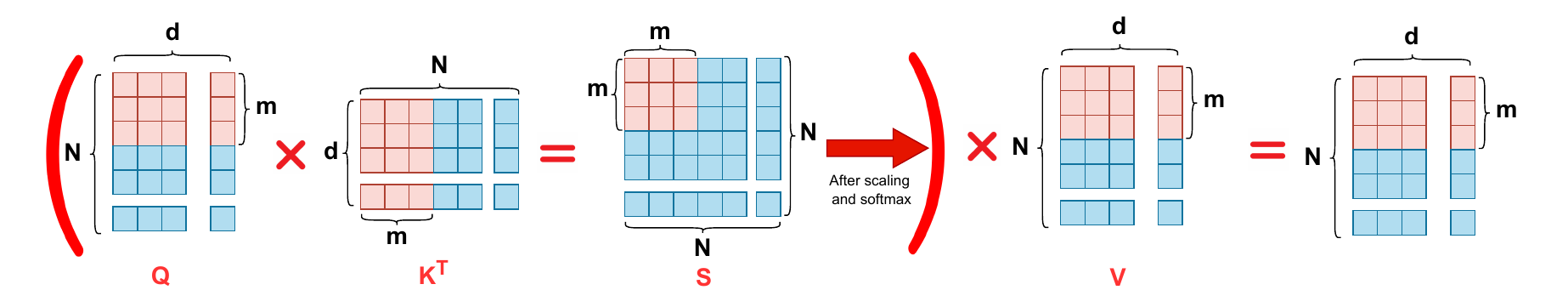}
\caption{Regular Scaled Dot-Product Attention mechanism. Here, N denotes the sequence length, d is the embedding dimension, and m represents the maximum number of tokens that can be stored as an $m\times d$ matrix within a single memory bank. Parts highlighted in red indicate portions stored locally within the same bank.}
\label{fig:reg_sdpa}
\end{figure*}

One baseline implementation of the attention mechanism in PIM architectures is to directly apply the standard Scaled Dot-Product Attention mechanism (SDPA), as shown in Figure~\ref{fig:reg_sdpa}. For load balancing, tokens are distributed across memory banks, where each bank stores $m$ tokens, with each token represented as a $1 \times d$ vector. We assume that the weight matrices ($W_Q$, $W_K$, $W_V$) are locally stored in each bank, following previous works such as \cite{zhou2022transpim}, as illustrated in Figure~\ref{fig:mapping}. Later in Section~\ref{scalability}, we further analyze the scalability implications of this setup.

Each bank computes the Query, Key, and Value vectors ($Q$, $K$, $V$) for its $m$ local tokens using the stored weight matrices. Thus, after this step, each bank holds $m$ computed Q/K/V vectors, as shown in Figure~\ref{fig:mapping}. The attention score computation will be discussed in detail in Section~\ref{multiplication}.

As depicted in Figure~\ref{fig:reg_sdpa}, the first step in attention calculation is multiplying $Q$ by $K^T$. For illustration, consider $m$ rows of the $Q$ and $K$ matrices that are stored in the same bank (highlighted in red). Each row of $Q$ must be multiplied by every column of $K^T$, which involves both local and non-local computations. For instance, multiplying the first row of $Q$ with the $m$ columns of $K^T$ that reside in the same bank can be performed locally. However, for the remaining columns, the $Q$ vector must be transferred from its source bank to the destination bank that holds the corresponding $K$ vectors. Each $Q$ vector needs to be transferred to $\frac{N}{m}$ banks, where $N$ is the total sequence length.
This implies that computing $QK^T$ requires $\frac{N}{m}$ movements per $Q$ vector, resulting in a total of $\frac{N^2}{m}$ vector movements (each of size $1 \times d$). Additionally, note that the resulting attention map $S$ is of size $N \times N$, and cannot fit entirely in a single bank. When each bank stores a $m \times m$ submatrix of $S$, the storage overflow requires further movement of these partial $S$ matrices to new banks, which introduces additional movement not captured in this estimate but accounted for in our evaluation.

Each bank now holds a $m \times m$ portion of the attention map $S$. As shown in Equation~\ref{eq:sda}, the scaling operation is applied locally. However, the softmax operation requires computing the sum of exponentials for each row of $S$. Since only $m$ elements of each row are stored in a single bank, and each full row spans across $\frac{N}{m}$ banks, the computation must involve inter-bank communication.

After computing $\exp(x)$ locally for each value in $S$, these partial results need to be aggregated across $\frac{N}{m}$ banks and then broadcast back for normalization. This results in $\frac{N}{m}$ movements per row and a total of $\frac{N^2}{m}$ movements. Even with optimization—such as sending aggregated sums of $m$ values at once—the total movement would still be $\frac{N^2}{m^2}$.

The final step involves multiplying the attention map $S$ by the Value matrix $V$. This operation, like the $QK^T$ step, includes both local and non-local computation. For instance, multiplying the first row of $S$ with the corresponding $m$ columns of $V$ can be performed locally, assuming they are co-located in the same bank. However, if parts of matrix $S$ were already moved to other banks due to storage limitations, additional data movement will be required even for these nominally local computations. However, for the remaining parts of the row, data must again move across $\frac{N}{m}$ banks to align with the relevant columns of $V$. Thus, the total movement is again $\frac{N^2}{m}$ or optimally $\frac{N^2}{m^2}$. Also, since not all intermediate results can be stored in the same bank due to limited storage, additional data movement is required (included in our evaluation but not in the above rough estimate).

Finally, all these intermediate products must be aggregated to generate the final output of the attention layer. Because the results are distributed across multiple banks, additional data movement (up to $\frac{N^2}{m^2}$) is incurred during accumulation.
Note that all these movement estimates are rough approximations and do not account for factors such as physical bank distance, contention, or memory interconnect architecture. However, our evaluation (Section~\ref{sec:evaluation}) incorporates these details for a precise and realistic analysis.

In summary, this baseline approach using standard SDPA in a PIM system incurs $O\left(\frac{N^2}{m}\right)$ data movement, making it inefficient and non-scalable for long sequences or high-throughput workloads.

\subsection{\paper{} Implementation of Attention in PIM Architecture}
\label{att_impl}

\paper{} adopts an alternative attention algorithm that significantly reduces data movement. Instead of explicitly computing the full $QK^T$ attention matrix, the algorithm first applies the softmax function separately to $Q$ and $K^T$. It then multiplies $\sigma(K^T)$ with the Value matrix $V$, followed by a multiplication with $\sigma(Q)$ to produce the final result, where $\sigma$ denotes the softmax operation.

This algorithm, introduced by Zhuoran et al. \cite{shen2021efficient}, was originally designed to improve computational efficiency. However, it does not explicitly address the critical bottleneck of data movement in memory—one of the main contributors to latency in LLMs. In this work, we adapt and extend their method for in-memory execution. Our version restructures the attention computation to reduce inter-bank communication and improve data locality, making it highly suitable for PIM architectures.

\begin{figure*}[t]  
    \centering
   \includegraphics[width=\textwidth]{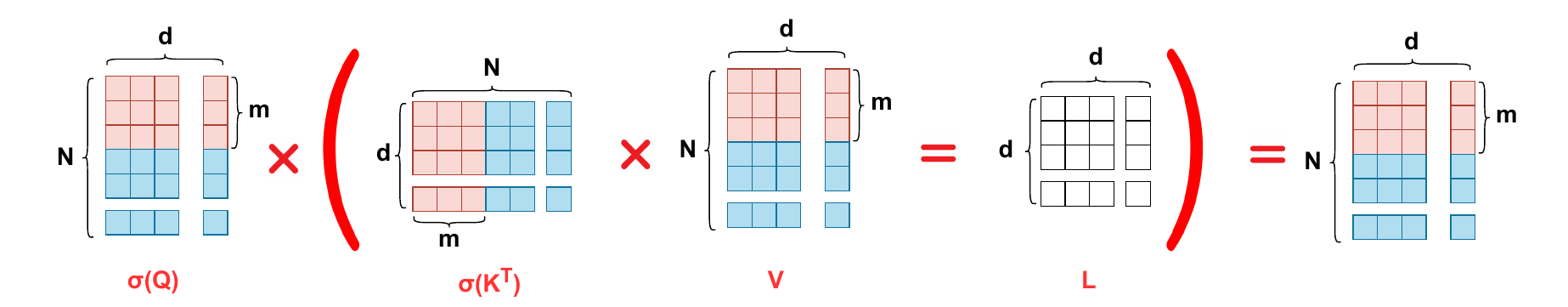}
\caption{\paper{} Scaled Dot-Product Attention. Parts highlighted in red indicate portions stored locally within the same bank. $\sigma$ denotes the softmax operation.}
\label{fig:ea_sdpa}
\end{figure*}

The first step in \paper{} is to compute $\sigma(Q)$ and $\sigma(K^T)$. For the softmax operation, we only need to calculate the $\sum \exp(x)$ for each row, where $x$ represents the values in that row. Computing $\sigma(Q)$ is entirely local because each row of $Q$ corresponds to a single query vector, which is fully stored within one bank (as illustrated in Figure~\ref{fig:ea_sdpa}).

However, computing $\sigma(K^T)$ involves some data movement. For each row of $K^T$, only $m$ values are stored locally in a bank, and the remaining values are distributed across the other $\frac{N}{m}$ banks. As a result, partial sums from these banks need to be transferred and accumulated, leading to $\frac{N}{m}$ data movements.

Next, we compute the multiplication $\sigma(K^T) \cdot V$. Notably, the $i$-th block of $m$ columns in $\sigma(K^T)$ is stored in the same bank as the corresponding $m$ rows of $V$ as shown in Figure~\ref{fig:ea_sdpa}, allowing this portion of the multiplication to be done locally. For example, when multiplying the first row of $\sigma(K^T)$ to $V$, its values are distributed across $\frac{N}{m}$ banks. Each of these banks holds $m$ elements from the row and the corresponding $m$ rows from $V$. The result of each local multiplication is a $1 \times d$ vector, and doing this across all rows produces a local $d \times d$ matrix in each bank.

These local matrices are then aggregated across banks to form the final result of $\sigma(K^T) \cdot V$. Since intermediate results are distributed across $\frac{N}{m}$ banks, the accumulation step also incurs $\frac{N}{m}$ data movements.

We refer to the resulting matrix as $L$, which is a compact $d \times d$ matrix, significantly smaller than the $N \times N$ attention matrix $S$ used in the standard attention approach (as shown in Figure~\ref{fig:reg_sdpa}).

Finally, $L$ must be multiplied by $\sigma(Q)$. In the worst-case scenario, if $L$ is stored in a single bank without replication, each row of $\sigma(Q)$ must be transferred to that bank. Since $\sigma(Q)$ is distributed across $\frac{N}{m}$ banks, this step requires up to $\frac{N}{m}$ data movements. 

As mentioned earlier, these estimations do not take into account the physical distance between memory banks or the specific characteristics of the bank architecture. Additionally, they do not reflect any architectural optimizations we introduce later in this paper, which are designed to further reduce data movement.

Overall, this approach reduces data movement to $O\left(\frac{N}{m}\right)$, which is significantly lower than that of the conventional implementation. It also requires substantially less memory capacity, as the matrix $L$ is much smaller than the matrix $S$.

\subsection{Architecture Design in \paper{}}
In this section, we present the architectural details of our proposed \paper{} design.

\subsubsection{Memory Architecture}
We adopt FIMDRAM \cite{kwon202125} as our baseline PIM architecture to leverage in-bank processing capabilities for computation. Each DRAM bank is equipped with local compute logic, enabling vector multiplications to be performed directly within the memory banks.

\subsubsection{In-bank Multiplication}
\label{multiplication}
As discussed in the previous sections, a large number of vector multiplications are executed locally within each bank. As an example, we provide the pseudocode for multiplying vectors $x$ and $y$ in Algorithm~\ref{algo:multiply}.

\begin{algorithm}
\caption{FP16 Dot Product of Vectors $x$ and $y$ (length $d$)}
\begin{algorithmic}[1]
\State \textbf{MOVE} SRF\_M, 0 \Comment{Clear accumulator}
\For{chunk = 0 to $(d / 128) - 1$}
    \Comment{Process 128 FP16 values per chunk (64 from $x$, 64 from $y$)}
    
    \For{$i = 0$ to $3$}
        \State \textbf{FILL} DRAM[$x$ + chunk $\times$ 128 + $i \times 16$] $\rightarrow$ GRF[$2i$]
        \State \textbf{FILL} DRAM[$y$ + chunk $\times$ 128 + $i \times 16$] $\rightarrow$ GRF[$2i + 1$] 
    \EndFor

    \For{$i = 0$ to $3$}
        \For{$j = 0$ to $15$}
            \State \textbf{FP16\_MAC} GRF[$2i$][$j$], GRF[$2i+1$][$j$], SRF\_M 
        \EndFor
    \EndFor
\EndFor
\State \textbf{MOVE} SRF\_M $\rightarrow$ GRF \Comment{Move final result to GRF}
\State \textbf{MOVE} GRF $\rightarrow$ DRAM \Comment{Store final result to memory}
\end{algorithmic}
\label{algo:multiply}
\end{algorithm}

\subsubsection{Softmax}
For softmax computation on a vector $x$, we need to calculate:

\begin{equation}
    \sigma(x_i) = \frac{\exp(x_i)}{\sum_j \exp(x_j)}
\end{equation}

Thus, computing $\exp(x)$ efficiently is critical. There are two common approaches for this calculation: a simple lookup table (LUT) \cite{yang2020retransformer,han2024sal}, and a Taylor series expansion \cite{zhou2022transpim}. 

The LUT approach requires a large table to store precomputed values and is restricted to a limited range of $x$. The Taylor series method is accurate near zero but suffers from a significant decline in accuracy as \( x \) deviates from zero. Moreover, it incurs higher computational complexity.

To overcome these limitations, we designed a hardware-friendly implementation that maintains both high accuracy and low complexity across a wider input range.

Our method is based on the identity:

\begin{equation}
e^x = 2^\alpha = 2^{\lfloor \alpha \rfloor + \{\alpha\}}, \quad \text{where} \quad \alpha = x \cdot \log_2 e
\end{equation}

Here, $\log_2 e$ is a constant, so computing $\alpha$ is fast. The integer part, $2^{\lfloor \alpha \rfloor}$, can be computed efficiently via bit shifting. The fractional part, $2^{\{\alpha\}}$, lies in the range $[0,1]$ and is approximated using a small LUT with linear interpolation. This approach significantly reduces memory usage and improves performance while retaining good accuracy.

To evaluate the effectiveness of our method, we conducted experiments using 10 million random values in the range \([-200, 0]\). Negative range is used, since frameworks like PyTorch \cite{ketkar2021introduction} first subtract the maximum value in the vector before applying the exponential function. We compared the average relative error of our method against both a simple LUT and the Taylor series expansion.

Our method, using only 0.5 KB of LUT storage, achieves an average relative error of approximately \(2.5 \times 10^{-3}\), which is comparable to the simple LUT method that uses 312.5 KB (625× larger table). In contrast, the Taylor series method yields a significantly higher average relative error of \(2.6 \times 10^{77}\).

We also evaluated all three methods on 10 million random numbers in the range \([-1, 0]\), where values are close to zero. In this case, the Taylor series achieves better accuracy, with an error of about \(5.9 \times 10^{-4}\), while the performance of our method and the LUT method remains consistent with previous results.

These experiments demonstrate that our proposed implementation offers competitive accuracy compared to existing methods, while significantly reducing both memory usage and computational complexity.

After computing each $\exp(x_i)$, the next step is to accumulate the results. If the values are stored locally, this accumulation can also be performed locally without requiring any data movement. However, for non-local values, a hierarchical accumulation approach is used, as described in Section~\ref{hierarchy}.



\subsubsection{Network}

For inter-bank communication, we adopt a network topology as shown in Figure~\ref{fig:mapping}, inspired by the design in \cite{rezaei2020nom}. In this architecture, each channel is composed of multiple banks, and inter-bank data communication is supported through horizontal and diagonal links. There is also links between channels.

To facilitate communication, a small buffer is included in each bank to temporarily store data being sent or received over the interconnect. These buffers help in decoupling the communication latency from computation.

Furthermore, a local controller (Ctrl) in each bank is responsible for routing the data. It uses a multiplexer to select which data should be forwarded to the arrays. 

\subsubsection{Hierarchical Transfer}
\label{hierarchy}

A naive approach to data transfer involves sending all data directly to a centralized unit for processing. Although conceptually simple, this method suffers from significant drawbacks, most notably high latency and congestion. Centralizing all communication creates a bottleneck, as data from multiple sources must traverse shared links, leading to increased traffic and longer delays.

To address this, \paper{} adopts a more scalable strategy called hierarchical transfer, where computation and communication are distributed across multiple levels and executed in parallel. In our design, this method is used for operations such as accumulation during matrix multiplication and softmax computation.

In the first step, within each bank group, accumulation is performed locally. For example, in each bankgroup, Bank 0 and Bank 1, sharing the same PCU, accumulate their results in parallel with Bank 2 and Bank 3. After this step, the results from each pair are further accumulated, e.g., the partial sum from Bank 1 is sent to Bank 3 for the next level of accumulation. This calculation under bankgroup is parallel among all bankgroups, so at the end each of them has a fully accumulated value stored in a designated bank (denoted as $b_{ij}$, where $i$ is the channel index and $j$ is the bank group index).

In the second step, this process continues in parallel across all channels. Specifically, the designated accumulation banks (e.g., $b_{i}$ from each bank group within a channel) exchange data with their neighboring bank groups, and further accumulation is performed. Since each channel consists of four bank groups, an additional level of transfer and accumulation is required to combine the intermediate results. After this final stage, the complete accumulated result for each channel is stored in a single designated bank, denoted as $b_i$.

If accumulation across multiple channels or dies is required, the same hierarchical mechanism is extended across dies.  

\begin{figure}[t]
\centering
\includegraphics[width=0.48\textwidth]{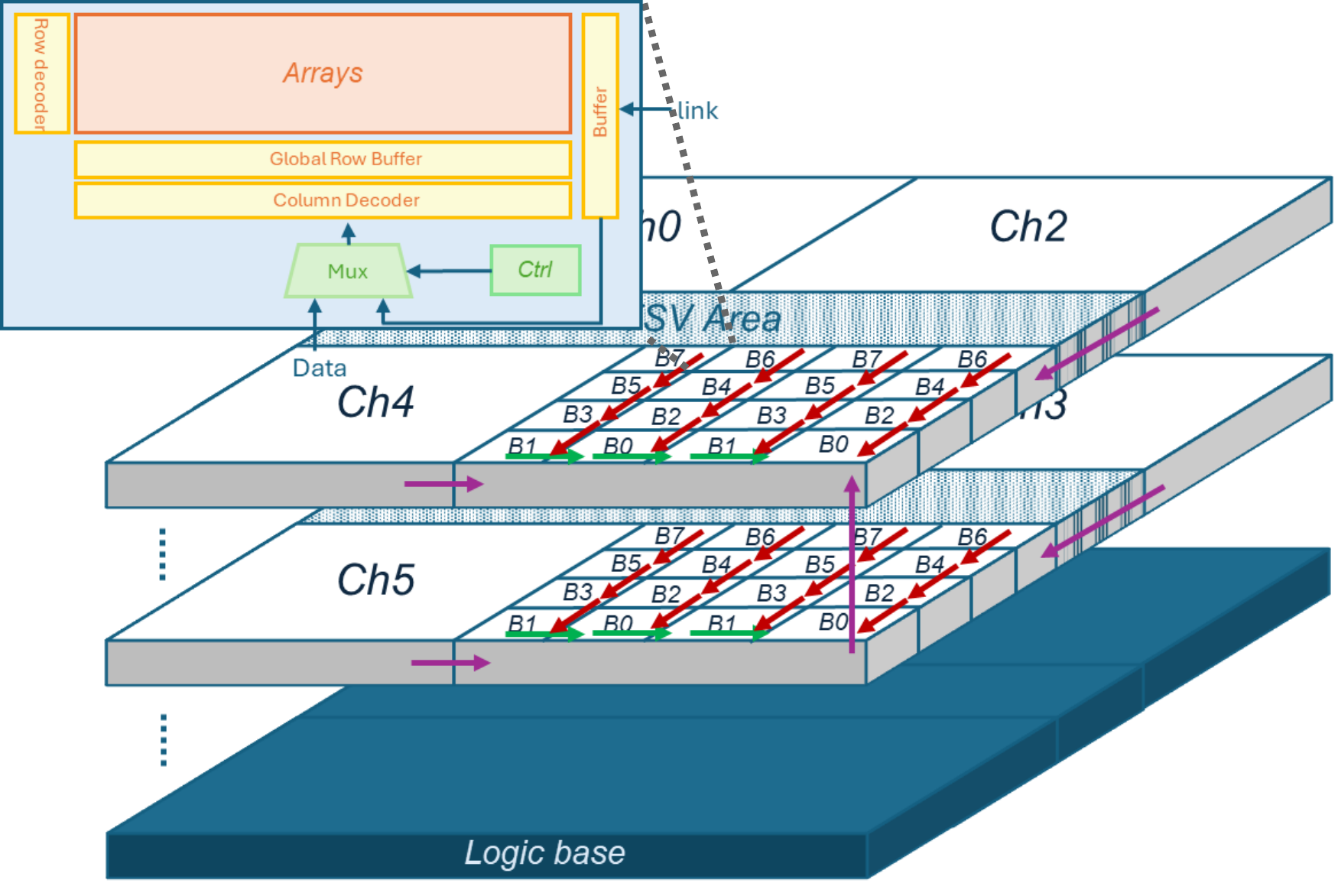}
\caption{Hierarchical data transfer architecture. Each colored arrow represents a different level of transfer: red for bank group level, green for channel level, and purple for die level.}
\label{fig:network}
\end{figure}

Figure~\ref{fig:network} illustrates the hierarchical transfer process using different colors: red arrows represent transfers at the bank group level, green arrows indicate transfers at the channel level, and purple arrows correspond to transfers at the die level.

Compared to the naive approach, this method offers several key advantages:

\begin{itemize}
    \item Reduced communication overhead: By performing early local and group-wise aggregation, the amount of interconnect traffic is significantly reduced.
    \item Lower latency: Parallel accumulation at each stage allows different parts of the memory hierarchy to work concurrently, avoiding the bottlenecks associated with serialized summation.
    \item Scalability: The structured, multi-level design naturally scales with the number of banks and channels, maintaining efficiency even in large memory configurations.
\end{itemize}

Overall, this hierarchical mechanism accelerates the computation, significantly reducing communication overhead, overall latency, and traffic.

\subsection{Scalability}
\label{scalability}

In the previous sections, we assumed (for simplicity of explanation) that the weight matrices  $W_K$, $W_Q$, and $W_V$, each of size $d \times d$, could fit entirely within a single bank. This assumption holds only for models with relatively small parameter sizes. However, \paper{} is designed to also handle cases where these weight matrices are too large to fit in a single bank, unlike prior approaches.

In this scenario, as before, we distribute the tokens across banks, such that each bank holds at most $m$ tokens. The weight matrices are then \emph{tiled column-wise}, where each bank holds a tile of $k$ columns. As a result, each weight matrix is distributed across $\frac{d}{k}$ banks, and the tokens are distributed across $\frac{N}{m}$ banks.

To compute the $k$, $Q$, and $V$ matrices, we move the weight tiles to the banks where the tokens reside. Specifically, each bank storing tokens must receive all column vectors from the $\frac{d}{k}$ weight banks. This requires data movement of $d$ column vectors to each of the $\frac{N}{m}$ token banks for each of the three matrices. Therefore, the total data movement will be $\frac{3Nd^2}{m}$.

\section{Evaluation Methodology and Results}
\label{sec:evaluation}

In this section, we present a comprehensive evaluation of \paper{} using simulation of several workloads.

\subsection{Workloads and Datasets}  
We evaluated \paper{} on a diverse set of benchmarks, including the GLUE benchmark suite \cite{wang2018glue}, which contains datasets designed to assess a model’s performance across various natural language understanding tasks. The datasets used in our evaluation are described below: 

\begin{itemize}  
    \item \textbf{MRPC (Microsoft Research Paraphrase Corpus):}  
    The MRPC dataset \cite{dolan2005automatically} consists of sentence pairs extracted from online news sources, annotated to indicate whether the sentences in each pair are semantically equivalent. This dataset evaluates a model’s ability to determine semantic similarity between sentence pairs.  

    \item \textbf{QQP (Quora Question Pairs):}  
    The QQP dataset \cite{wang2018glue} comprises pairs of questions sourced from the Quora platform. The objective of this task is to identify whether two questions are semantically equivalent.  

    \item \textbf{STS-B (Semantic Textual Similarity Benchmark):}  
    STS-B \cite{cer2017semeval} is a dataset that evaluates a model’s ability to predict the semantic similarity between sentence pairs. Each pair is annotated with a similarity score ranging from 1 to 5, representing the degree of semantic equivalence.  

    \item \textbf{MNLI (Multi-Genre Natural Language Inference):}  
    The MNLI dataset \cite{williams2017broad, wang2018glue} is a large-scale corpus for textual entailment tasks. It includes sentence pairs from various genres annotated with entailment labels (entailment, contradiction, or neutral). We used both the matched (in-domain) and mismatched (cross-domain) validation splits, containing 19,647 pairs each.  

    \item \textbf{QNLI (Question Natural Language Inference):}  
    QNLI \cite{rajpurkar2016squad, wang2018glue} is a binary classification task derived from the Stanford Question Answering Dataset (SQuAD). The goal is to determine if a given context sentence answers a corresponding question.  

    \item \textbf{RTE (Recognizing Textual Entailment):}  
    The RTE dataset \cite{bentivogli2009fifth, wang2018glue} evaluates textual entailment, requiring models to predict whether a hypothesis logically follows from a given premise.  

    \item \textbf{WNLI (Winograd NLI):}  
    The WNLI dataset \cite{levesque2012winograd, wang2018glue} is adapted from the Winograd Schema Challenge. It presents challenging reasoning tasks where the model must determine whether a hypothesis follows logically from a given premise.  

    \item \textbf{CoLA (Corpus of Linguistic Acceptability):}  
    The CoLA dataset \cite{warstadt2019neural} assesses a model’s ability to distinguish between grammatically acceptable and unacceptable sentences.  

    \item \textbf{SST-2 (Stanford Sentiment Treebank):}  
    The SST-2 dataset \cite{socher2013recursive} contains sentences from movie reviews annotated with sentiment labels. The task is to classify each sentence as expressing positive or negative sentiment.  
\end{itemize}

We also utilized additional datasets featuring longer contexts, including the \textbf{PubMed} and \textbf{ArXiv} datasets \cite{cohan2018discourse}, which contain scientific papers from the biomedical and general scientific domains, respectively. These datasets are valuable for a wide range of tasks, such as information retrieval, document classification, and other advanced natural language processing applications. Additionally, we incorporated \textbf{GovReport} \cite{govreport}, which comprises reports authored by government research agencies, and \textbf{WikiHop} \cite{wikihop}, a dataset consisting of paragraphs from Wikipedia. Lastly, we utilized the \textbf{IMDB} dataset \cite{imdb}, providing movie reviews that serve as a benchmark for sentiment analysis and text classification tasks. 

\subsection{Experimental Setup}

We modify the DAMOV simulation framework~\cite{oliveira2021damov} to implement and evaluate our proposed architecture. DAMOV seamlessly integrates the ZSim~\cite{zsim} CPU simulator and the Ramulator~\cite{ramulator} memory simulator. This framework allows us to simulate a configurable number of traditional CPU cores or PIM cores with various memory technologies, including HBM. In our experiments, we simulate our \paper{} architecture on top of FIMDRAM (i.e., HBM-PIM) memory with 8 channels. Additional system configurations for \paper{} are detailed in Table~\ref{table:config}. 

To evaluate our method in Section~\ref{pref_analysis}, except for the accuracy evaluation, we conducted experiments on a standard transformer block consisting of MHA, FFN, and Add \& Normalization layers, as shown in Figure~\ref{fig:transformer}. This block serves as a fundamental component in transformer architectures.

For the accuracy evaluation (presented in Section~\ref{pref_analysis}), we used the \texttt{bert-base-uncased} model and replaced all self-attention layers with our proposed implementation.

\begin{table}[t]
    \centering
    \renewcommand{\arraystretch}{1.2} 
    \setlength{\tabcolsep}{8pt} 
    \caption{Architectural Parameters of \paper{}.}
    \label{table:config}
    \begin{tabular}{|c|c|} 
        \hline
        \textbf{Parameter} & \textbf{Value} \\ 
        \hline
        Num of channels & 8 \\ \hline
        Num of banks per channel  & 16 \\ \hline
        Num of PCUs per channel & 8 \\ \hline
        Num of Multipliers per PCU & 16 of FP16 Mult \\ \hline
        Num of Adders per PCU & 16 of FP16 Add \\ \hline
        Processing operation speed & 300MHz \\ \hline
        Total size & 4GB \\ \hline
        \multirow{4}{*}{\centering Memory Timing} & $t_{RC}=24$, $t_{RAS}=17,$ \\  
        & $t_{RRD}=5$, $t_{RCD}=7,$ \\  
        & $t_{CL}=7$, $t_{WR}=8,$ \\  
        & $t_{CCD}=2$ \\  
        \hline
    \end{tabular}
    \vspace{-4pt}
\end{table}

\subsection{Baseline}
As the baseline, we consider the standard transformer implementation described in Section~\ref{base_impl} and illustrated in Figure~\ref{fig:reg_sdpa}, using FIM-DRAM for in-memory computation. For the softmax operation, we use a Taylor-series approximation, and for inter-bank communication, the naive broadcast mechanism is employed, as discussed in Section~\ref{hierarchy}. The HBM structure is kept consistent with our optimized implementation to ensure a fair comparison.

This design represents a straightforward in-memory implementation without any specific optimizations for data movement or bank-level parallelism.

\subsection{Performance Analysis}
\label{pref_analysis}

\begin{figure}[t]
\centering
\includegraphics[width=0.49\textwidth]{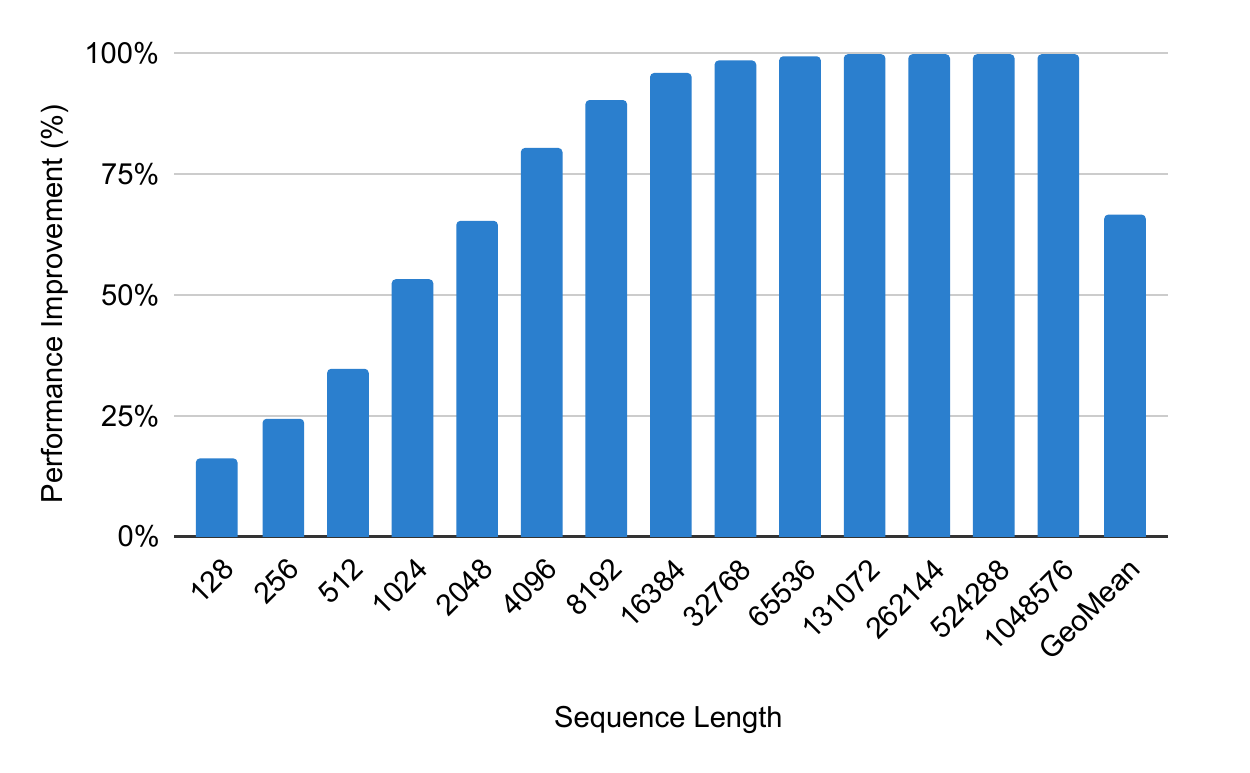}
\vspace{-22pt}
\caption{Execution time improvement vs. baseline for sequence lengths ranging from 128 to 1M.}
\label{fig:result_64core_128d}
\end{figure}

\noindent\textbf{Evaluation on different sequence lengths: } To evaluate the efficiency of our proposed method, we conducted experiments across a range of sequence lengths, from 128 tokens up to 1 million. Figure~\ref{fig:result_64core_128d} presents a performance comparison between \paper{} and the baseline. The results show that the performance gains of \paper{} increase with sequence length. Specifically, \paper{} achieves improvements ranging from 16.05\% to 99.99\%, with a geometric mean speedup of 66.42\% over the baseline. These results demonstrate that although \paper{} provides noticeable improvements even for shorter sequences, it is particularly effective for longer sequences, where the benefits of reduced data movement and improved parallelism become more significant.

\begin{figure}[t]
\centering
\includegraphics[width=0.48\textwidth]{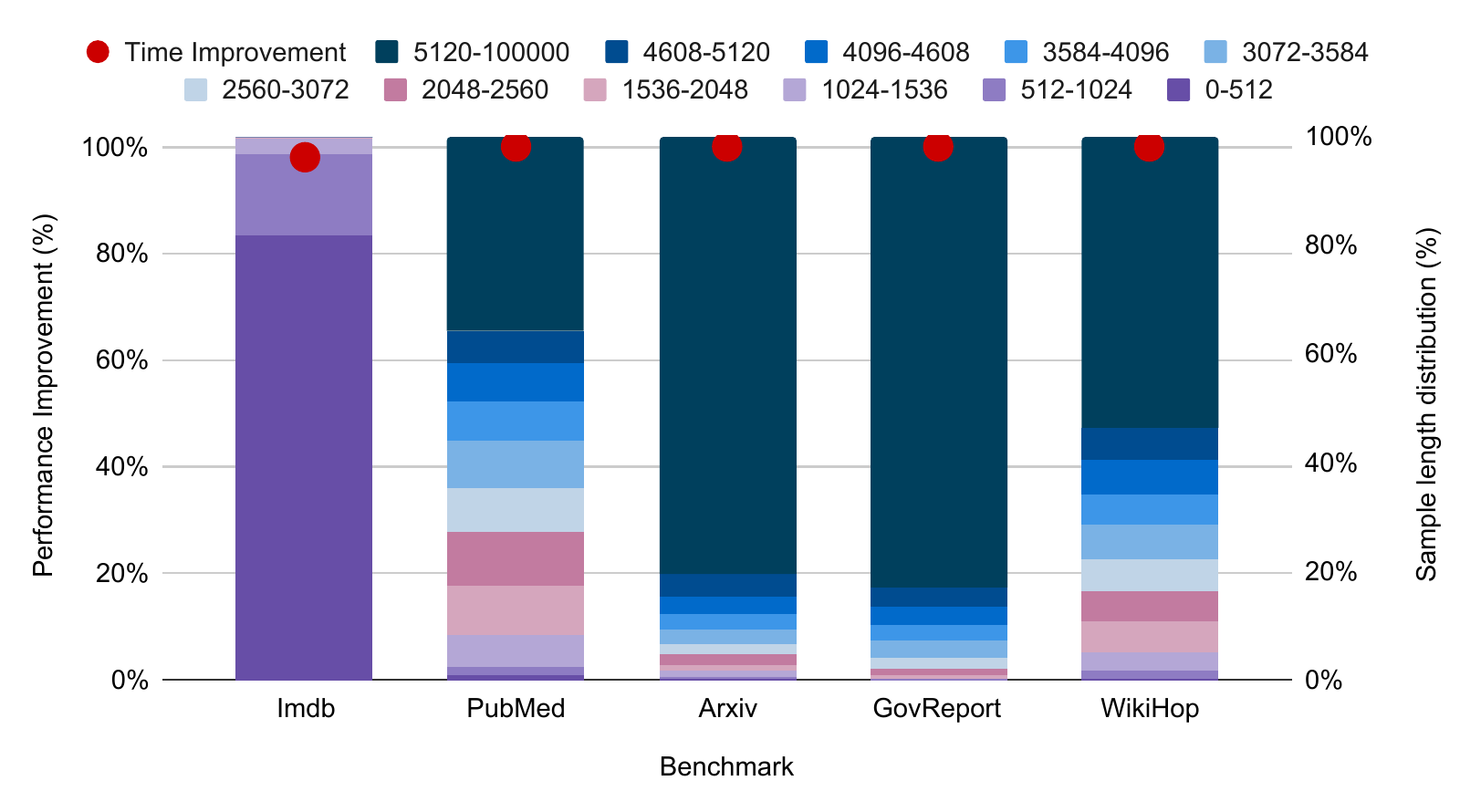}
\vspace{-10pt}
\caption{Execution time improvement vs. baseline across different datasets with longer documents (indicated by red points, left y-axis), along with the sequence length distribution for each dataset (depicted by the bars, right y-axis).}
\label{fig:large_res}
\vspace{-10pt}
\end{figure}

\noindent\textbf{Evaluation on long-context datasets:} We further evaluated our approach on five datasets that consist of long sequences, such as full documents and articles. Specifically, we used IMDB, PubMed, ArXiv, GovReport, and WikiHop. Figure~\ref{fig:large_res} illustrates the distribution of sequence lengths in these datasets, where darker blue shades represent longer sequences. As shown in the figure, the sequence length distributions vary significantly across datasets, for example, IMDB contains relatively shorter sequences. The figure also presents performance improvements, indicated by red dots. Our approach achieves a geometric mean performance improvement of 99.60\%, demonstrating a substantial reduction in execution time.

\begin{figure}[t]

\centering
\includegraphics[width=0.48\textwidth]{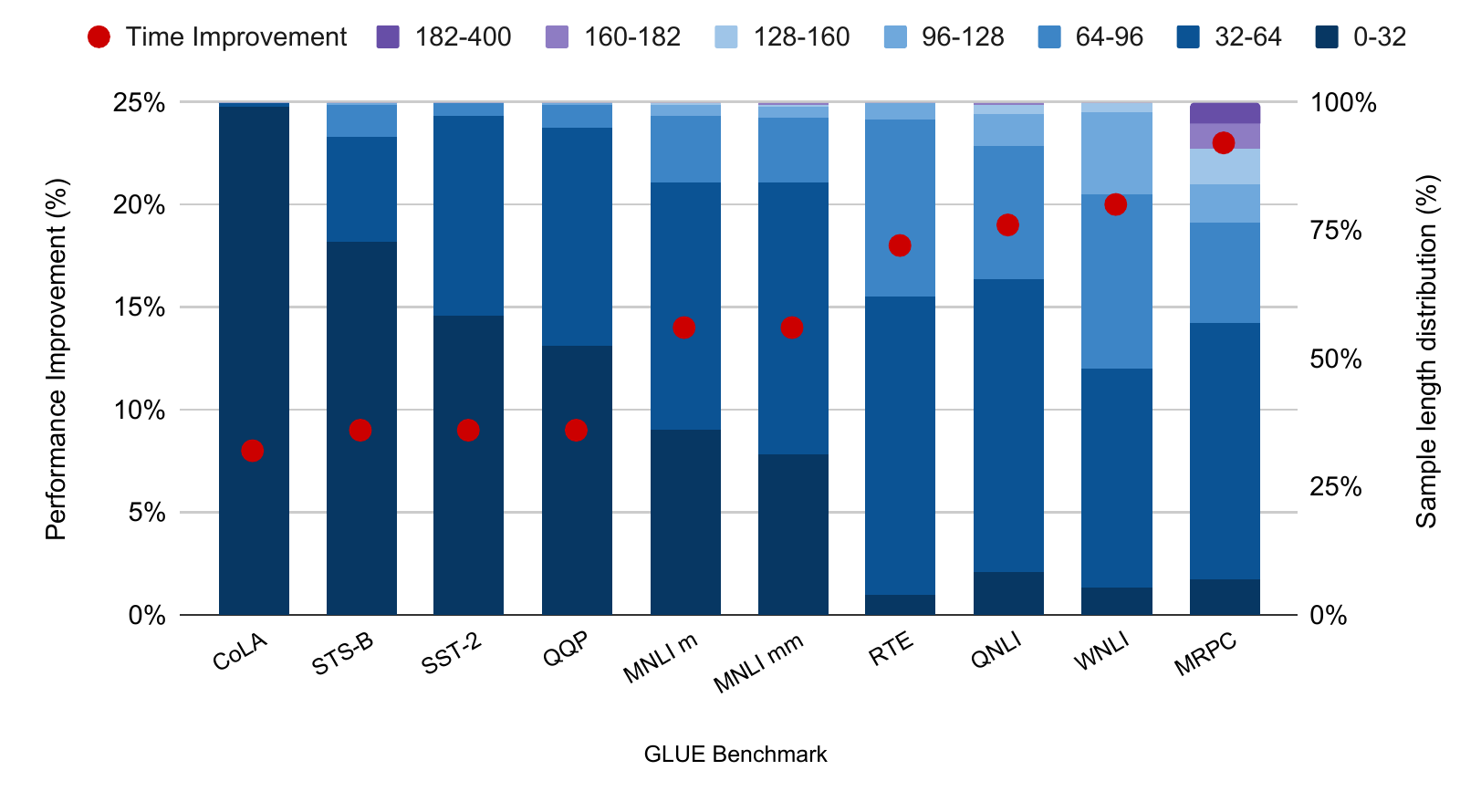}
\vspace{-10pt}
\caption{Execution time improvement on the GLUE benchmark for our method compared to baseline (represented by dots, left y-axis), and sequence length distribution in each dataset (represented by bars, right y-axis).}
\label{fig:glue_res}
\vspace{-10pt}
\end{figure}

\noindent\textbf{Evaluation on short-context datasets:} We also tested our approach on datasets with shorter sequence lengths to determine whether our method still provides benefits in such scenarios. We evaluated \paper{}'s performance on the GLUE benchmark, which consists of several natural language understanding tasks with varying sequence lengths. Figure~\ref{fig:glue_res} shows the distribution of sequence lengths for each dataset under GLUE separately. As this figure shows, datasets contain short sequences.  

This figure also demonstrate performance improvement by red dots.\paper{} achieved a performance improvement ranging from 8.47\% to 22.61\%, with a geometric mean improvement of 13.44\%. While the improvements are smaller compared to those achieved from longer sequences, the consistent performance gain highlights the generality of our approach across different sequence lengths.

Additionally, the distribution of sequence lengths across the GLUE datasets has a clear correlation with improvement: datasets with a higher proportion of longer sequences exhibit greater performance improvements. This is expected intuitively since data movement overhead of the baseline grows quadratically with sequence length, whereas our proposal mitigates this effect more effectively.

\textbf{Accuracy evaluation:} Since our algorithm approximates the standard Scaled Dot-Product Attention, it is important to evaluate its impact on model accuracy. On the GLUE benchmark, we observed an average performance drop of 0.58\%, while on IMDB, the drop was only 0.32\%. Interestingly, for GovReport, PubMed, and ArXiv, the model achieved performance improvements of 0.72\%, 4.29\%, and 1.37\%, respectively.

These results indicate that our approximation introduces negligible changes in accuracy, making it well-suited for efficient inference in transformer models.

\subsection{Comparison to Prior Transformer Accelerators}
We compared the performance improvement of \paper{} against previous transformer accelerators.

\begin{figure}[t]
\centering
\includegraphics[width=0.48\textwidth]{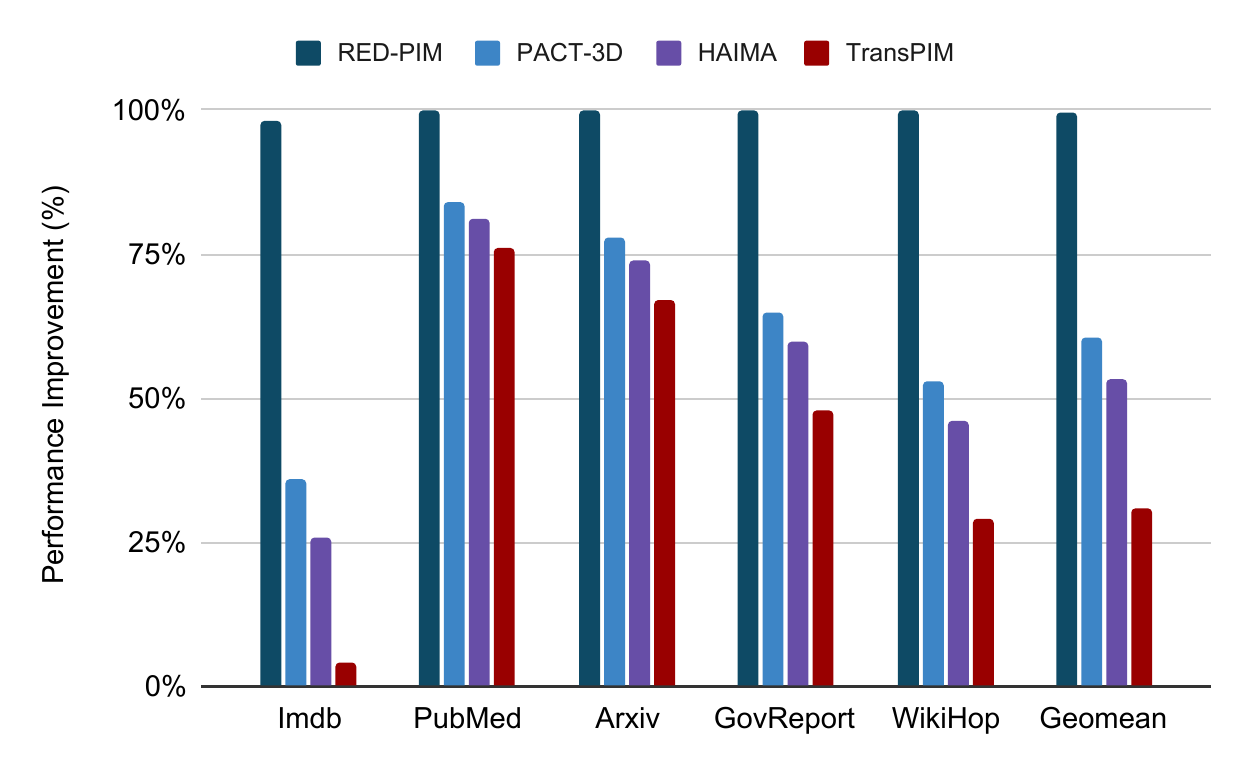}
\vspace{-10pt}
\caption{Performance improvement analysis of \paper{}, TransPIM \cite{zhou2022transpim}, HAIMA \cite{ding2023haima}, and PACT-3D \cite{singh2024dram} on datasets with larger documents (all improvements are relative to the baseline).}
\label{fig:large_compare}
\vspace{-10pt}
\end{figure}

\noindent\textbf{Comparison on long-context datasets:} Figure~\ref{fig:large_compare} presents the performance improvements on datasets with larger document sizes. \paper{} achieves the highest speedup, with a geometric mean improvement of 99.60\%, significantly outperforming all other methods. PACT-3D (60.53\%) and HAIMA (53.30\%) also deliver notable improvements, though their performance is lower on average than that of \paper{}. TransPIM, while effective in certain scenarios, shows only modest improvements on datasets such as IMDB.

\begin{figure}[t]
\centering
\includegraphics[width=0.48\textwidth]{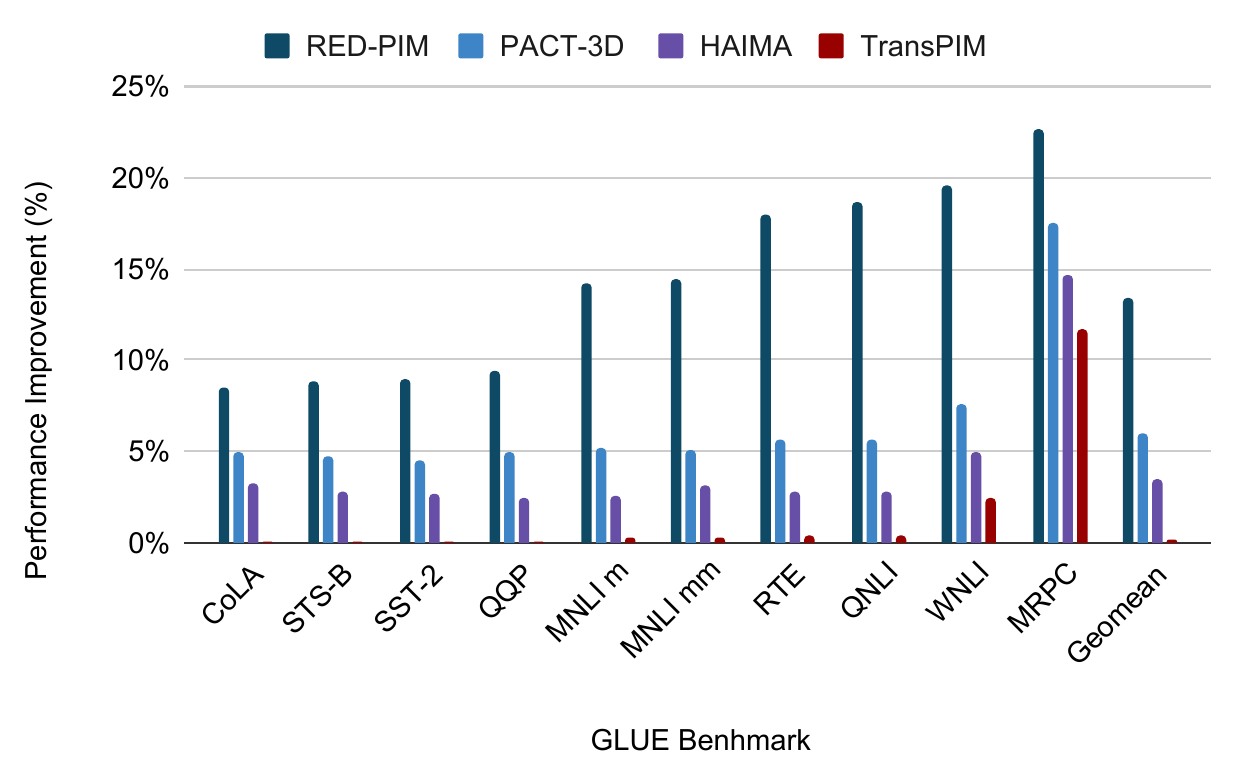}
\caption{Performance improvement comparison of \paper{}, TransPIM \cite{zhou2022transpim}, HAIMA \cite{ding2023haima}, and PACT-3D \cite{singh2024dram} on the GLUE benchmark (all performance improvements are relative to the baseline) .}
\label{fig:glue_compare}
\vspace{-10pt}
\end{figure}

\noindent\textbf{Comparison on short-context datasets:} Figure \ref{fig:glue_compare} illustrates performance gains on the GLUE benchmark, where \paper{} consistently outperforms all other methods across all tasks. TransPIM \cite{zhou2022transpim} achieves moderate improvements, particularly in MRPC and WNLI, while PACT-3D and HAIMA exhibit more modest gains across most benchmarks. The geometric mean results highlight \paper{} as the most effective approach, delivering an overall performance improvement of 13.44\%, compared to 5.98\% for PACT-3D, 3.47\% for HAIMA, and 0.15\% for TransPIM (all relative to the baseline).

\subsection{Sensitivity Analysis}

\subsubsection{Sensitivity to embedding size (\texorpdfstring{$d$}{d})}

In the previous experiments, we evaluated \paper{} using different input sequence lengths. Here, we analyze the effect of embedding size (\(d\)) on performance. The embedding size impacts both the amount of data transferred and the number of computations. For instance, as discussed in Section~\ref{att_impl}, each query vector that must be moved to the bank storing matrix \(L\) has a size of \(1 \times d\), so increasing \(d\) results in greater data movement. Additionally, the matrix \(L\) has dimensions \(d \times d\), meaning that a larger \(d\) leads to increased computational cost during attention calculation.

\begin{figure}[t]
\centering
\includegraphics[width=0.48\textwidth]{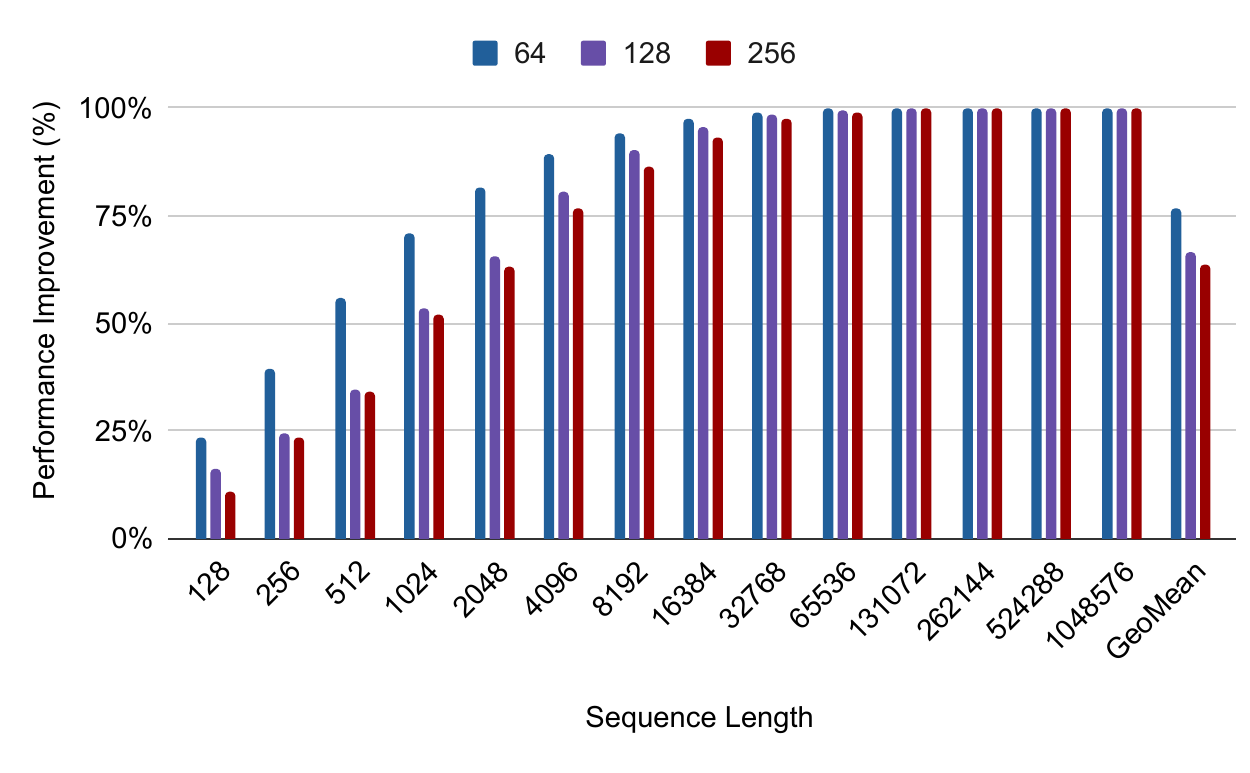}
\vspace{-10pt}
\caption{Sensitivity of model performance to embedding size (d) across varying sequence lengths, including the geometric mean performance aggregated over all sequence lengths.}
\label{fig:sens_d}
\vspace{-10pt}
\end{figure}

Figure~\ref{fig:sens_d} shows the performance improvement of \paper{} across varying sequence lengths for different embedding sizes (64, 128, and 256). The results indicate that performance generally improves with longer sequences, as they benefit more from our optimization strategy. Additionally, configurations with smaller \(d\) values show greater improvements due to reduced data transfer and lower computational cost.

The geometric mean results further suggest that our method remains effective across different embedding sizes, with the differences diminishing as sequence length increases. This is because, in longer sequences, the effect of the number of tokens (\(N\)) on number of data movement becomes dominant, thereby reducing the relative impact of \(d\).

\subsubsection{Sensitivity to the number of banks}

The number of memory banks per pseudo channel(sub channel) can also affect performance. Increasing the number of banks within a single die reduces inter-die communication overhead, which is generally more costly than intra-die communication. Therefore, having more banks per pseudo channel may lead to better performance compared to configurations with fewer banks and more dies.

\begin{figure}[t]
\centering
\includegraphics[width=0.48\textwidth]{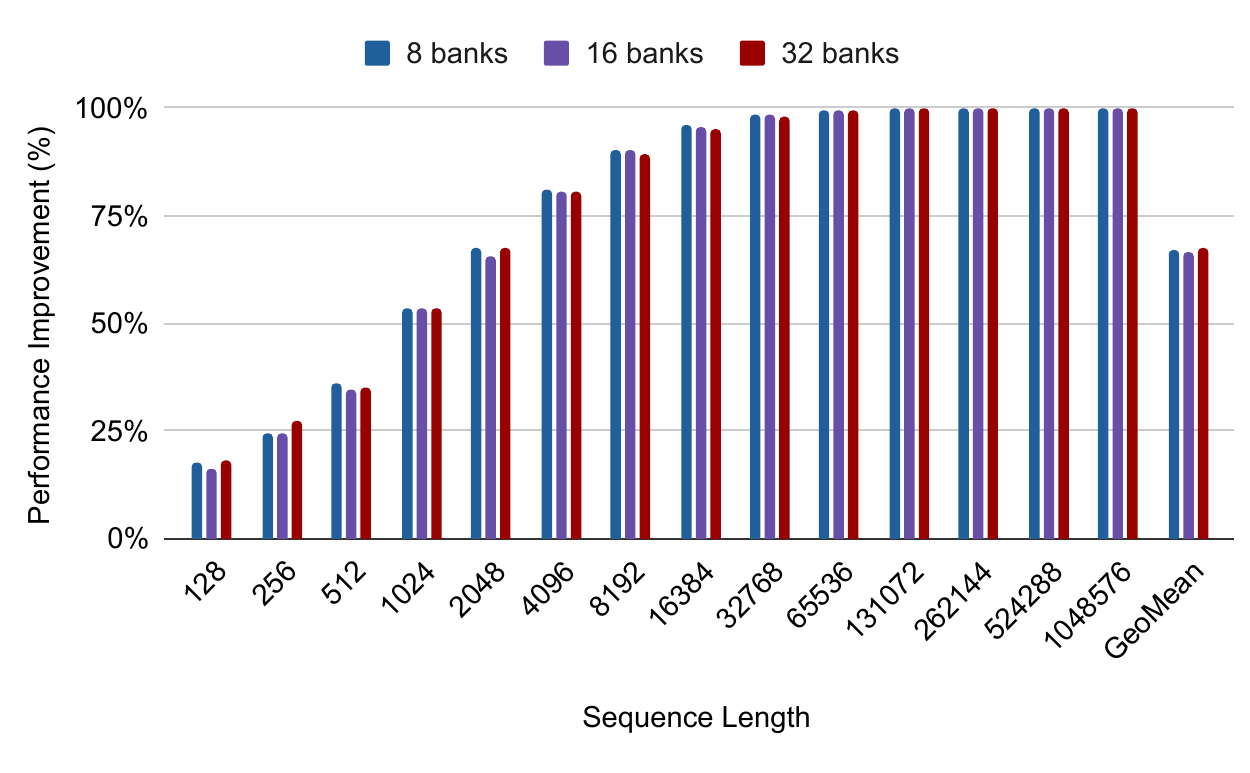}
\vspace{-10pt}
\caption{Sensitivity of model performance to varying numbers of memory banks.}
\label{fig:sens_bank}
\vspace{-10pt}
\end{figure}

Figure~\ref{fig:sens_bank} illustrates the impact of varying the number of memory banks (8, 16, and 32) on performance improvement across different sequence lengths. The results show a similar trend as in previous experiments: performance increases with longer sequences. The geometric mean results reveal a slight improvement when using 32 banks. Interestingly, the configuration with 16 banks shows slightly lower performance than the 8-bank setup. This behavior is attributed to interconnect bottlenecks: as the number of banks increases, simultaneous communication between them can saturate internal data paths or memory controller bandwidth. In such cases, the added parallelism is not fully utilized and instead leads to increased latency, offsetting the benefits of additional banks.

Overall, these findings demonstrate that \paper{} performs effectively across varying embedding sizes and memory bank configurations, ensuring robust performance under diverse workload characteristics.

\section{Related Work}
\label{related-work}

PIM implementations for transformers have attracted attention for reducing data movement and improving efficiency \cite{yang2020retransformer, zhou2022transpim, ding2023haima, singh2024dram, sridharan2023x, laguna2022hardware, kang2021framework}. Early work like ReTransformer \cite{yang2020retransformer} used ReRAM-based vector-matrix and matrix-matrix multiplication and lookup-based softmax to accelerate QKV and attention computations. However, such designs suffer from low precision and reliability, as well as limitations of ReRAM technology, which faces practical obstacles to large-scale manufacturing and has not yet been commercialized.

TransPIM \cite{zhou2022transpim} uses an HBM-based architecture that distributes tokens across banks and shifts keys in a Ring Broadcast and Compute pattern for attention. It employs Ambit \cite{seshadri2017ambit} for bit-serial point-wise multiplications and includes an Auxiliary Compute Unit with an adder tree and buffer in each bank. Exponentiation is approximated using a fifth-order Taylor expansion via PIM operations. While the ring pattern reduces data movement, serial processing introduces high latency, each 8-bit multiplication requires hundreds of cycles due to repeated Activate-Activate-Precharge (AAP) operations. AAP also disrupts DRAM timing and requires decoder changes for parallel row activation. As a result, performance gains are limited, particularly for long sequences.

HAIMA \cite{ding2023haima} optimizes computation by coordinating SRAM, DRAM, and the host. The host handles softmax and add-normalization, while other operations are distributed across memory hierarchies. However, frequent data transfers between SRAM, DRAM, and the host introduce high energy consumption and latency, with bus transactions becoming a performance bottleneck.

PACT-3D \cite{singh2024dram} employs NMP-Units with Neuron Processing Elements (NPEs) in each DRAM layer for in-memory computation, while RISC-V processors in the HBM logic layer handle normalization and softmax. However, integrating NPEs and RISC-V cores increases hardware overhead and chip complexity, and high-precision computations with NPEs result in longer latencies.  Additionally, frequent data movement between DRAM banks and the RISC-V processors introduces further latency and data movement overhead.

Moreover, there have been efforts to improve the computational efficiency of self-attention mechanisms using locality-sensitive hashing and bucketing methods. Notable examples include IMCAT \cite{laguna2021memory} and Cai et al \cite{cai2024memory}. These approaches reduce computation cost but increase complexity due to the bucketing and hashing operations, potentially increasing latency and memory overhead. Furthermore, IMCAT hashes keys into binary signatures and computes self-attention using only the $m$ closest keys by Hamming distance. This risks losing important information by ignoring many relevant token relationships.

In summary, while prior works have explored various Processing-in-Memory solutions for transformer models, they often target small-scale models or rely on impractical or overly complex hardware mechanisms. These designs struggle to scale to larger models with high-dimensional weights and long input sequences. Most importantly, they do not address the significant challenge of data movement.

\section{Conclusion}
\label{sec:conclusion}
In this work, we addressed a key bottleneck in transformer models, the excessive data movement incurred during self-attention operations, by proposing \paper{}, a PIM-based solution tailored for High-Bandwidth Memory (HBM) architectures. \paper{} restructures attention computation to operate on a compact intermediate representation, significantly reducing memory usage and inter-bank communication. Our algorithmic and architectural co-design leverages local computations within memory banks, efficient inter-bank communication strategies, and hardware-friendly softmax approximation to achieve scalable and low-latency transformer inference.

To evaluate our approach, we developed a simulation framework based on DAMOV, which enables detailed analysis of data movement in HBM architectures, showing up to a 99.99\% improvement for long sequences and an average improvement of 66.42\% over a simple PIM baseline with little impact on model accuracy.


Along with these improvements, some limitations remain to be explored in future work:

\begin{itemize}
    \item \textbf{Reduced Usable Bank Capacity:} The use of FIMDRAM involves allocating portions of each memory bank for PCUs, which reduces the available storage.

    \item \textbf{Simulation-Based Evaluation:} Due to the lack of available commercial or open-source hardware support, our PIM architecture is evaluated through simulation, which models data movement and in-memory computation.   

    \item \textbf{Focus on Inference:} Our work focuses on inference. Adapting \paper{} for training introduces challenges like gradient computation and intermediate data storage, which require further analysis and optimization.
\end{itemize}


\bibliographystyle{IEEEtranS}
\bibliography{refs}

\end{document}